\documentclass[letterpaper, 10 pt, conference]{ieeeconf}  

\usepackage{graphicx}
\usepackage{subfigure}
\usepackage{amsmath}
\usepackage{makecell}
\usepackage{multirow}
\usepackage{pifont}
\usepackage{amssymb}
\usepackage{amsmath}
\usepackage[utf8]{inputenc}

\IEEEoverridecommandlockouts                              

\title{\LARGE \bf
Accurate Visual-Inertial SLAM by Feature Re-identification
}

\author{Xiongfeng Peng$^{1}$, Zhihua Liu$^{1}$, Qiang Wang$^{1}$, Yun-Tae Kim$^{2}$, Myungjae Jeon$^{2}$
\thanks{Xiongfeng Peng$^{1}$, Zhihua Liu$^{1}$ and Qiang Wang$^{1}$ are with SAIT-China Lab, Samsung R\&D Institute China-Beijing, China
        {\tt\small \{xf.peng, zhihua.liu, qiang.w\}@samsung.com}}%
\thanks{Yun-Tae Kim$^{2}$ and Myungjae Jeon$^{2}$ with Multimedia Processing Lab, Samsung Advanced Institute of Technology, South Korea
        {\tt\small \{ytae.kim, myungje.jeon\}@samsung.com}}%
}

\begin{document}

\maketitle
\thispagestyle{empty}
\pagestyle{empty}

\begin{abstract}

We propose a novel feature re-identification method for real-time visual-inertial SLAM. The front-end module of the state-of-the-art visual-inertial SLAM methods (e.g. visual feature extraction and matching schemes) relies on feature tracks across image frames, which are easily broken in challenging scenarios, resulting in insufficient visual measurement and accumulated error in pose estimation. In this paper, we propose an efficient drift-less SLAM method by re-identifying existing features from a spatial-temporal sensitive sub-global map. The re-identified features over a long time span serve as augmented visual measurements and are incorporated into the optimization module which can gradually decrease the accumulative error in the long run, and further build a drift-less global map in the system. Extensive experiments show that our feature re-identification method is both effective and efficient. Specifically, when combining the feature re-identification with the state-of-the-art SLAM method \cite{VISLAM1}, our method achieves 67.3\% and 87.5\% absolute translation error reduction with only a small additional computational cost on two public SLAM benchmark DBs: EuRoC and TUM-VI respectively.
\end{abstract}

\section{INTRODUCTION}

Accurate 3D pose estimation of a moving camera is an important task in computer vision and has attracted more and more attention in recent years. It provides a fundamental function for many applications, such as augmented reality (AR) on smart phones or glasses, robot navigation and autonomous driving.

Visual-inertial simultaneous localization and mapping (VI-SLAM) is one of the most promising methods providing precise navigation in a 3D world through the fusion of cameras and inertial sensors. Comparing with the visual-inertial odometry (VIO) methods \cite{VIO1,VIO2,VIO3,VIO4,VIO5,VIO6,VIO7,VIO8,VIO9,VIO10}, VI-SLAM methods \cite{VISLAM1,VISLAM2,VISLAM3,VISLAM4,VISLAM8,VISLAM5,VISLAM6,VISLAM7} have the advantage of building a global map of 3D sparse features of the surrounding environment, which can effectively bound tracking drift.
Patrick G. et al. \cite{VISLAM2} propose a SchmidtEKF based VI-SLAM method which employs keyframe-aided 2D-2D feature matching to find reliable correspondences between current 2D visual measurements and 3D map features.
Qin T. et al. \cite{VISLAM3} propose a monocular visual-inertial SLAM which merges current map with previous map by loop detection and relocalizes camera.
Campos C. et al. \cite{VISLAM8} propose a multiple map system that relies on a new place recognition method with improved recall.
The above methods rely on feature matching between two images or image matching with similar poses, i.e. the baseline and orientation difference between two images cannot be very large so that direct matching based on feature descriptors such as Bag-of-Words \cite{c12} can yield reliable results.



There are also methods which try to establish feature correspondence in relatively larger viewpoint changes by multi-frame feature integration \cite{c5, c17}, or by 2D-3D feature matching between the current frame and a local map constructed from a set of neighboring keyframes on the co-visibility graph \cite{VISLAM6}. Typically, two neighboring co-visibility graph nodes share a number of common features obtained through either feature tracking or matching. However, such methods heavily rely on feature tracking or matching across image frames in temporal neighborhood and cannot build efficient visual measurements, which result in significant camera drift in challenging scenarios such as occlusion or large viewpoint changes.
\begin{figure}[t]
\begin{center}
\includegraphics[width=1.0\linewidth]{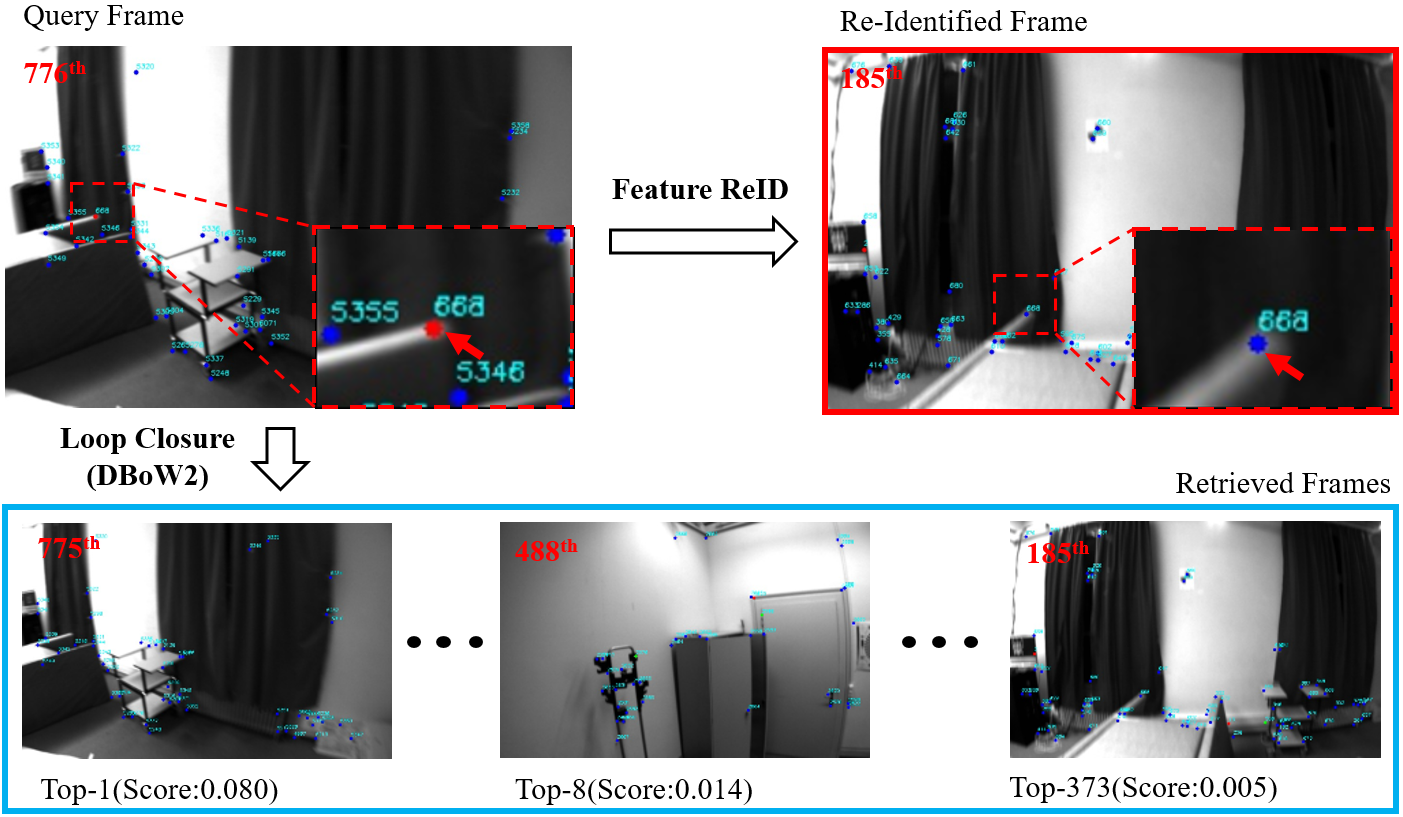}
\end{center}
\caption{The comparison of loop closure with our proposed feature re-identification method. For the query frame 776$^{th}$ on EuRoC DB, DBoW2 method is implemented to detect loop and the top-k retrieved frames are temporally close to the query frame (k=7). However, our proposed feature ReID method can successfully retrieve early frame 185$^{th}$ and build the constraints between them.}
\label{fig:net}
\end{figure}
To get drift-less camera pose with more efficient visual measurements, in this paper, we propose a novel visual inertial SLAM framework with feature re-identification (ReID) method. The method re-identifies new detected features in the current frame whose corresponding 3D map points have already been reconstructed previously. Note that the previous works on the loop closure detection also build the connection between the current frame and historical map/frames. However, our feature ReID method exploits a fused IMU-aided, geometry and appearance constraint which enables to re-identify reliable features in a long time span while loop closure methods only focus on neighboring co-visibility keyframes before the closed loop is found. Fig.1 represents a detailed comparison of our feature ReID and the loop closure method.

Our proposed feature ReID method in this paper is not straightforward since the features to be re-identified may be subjected to large viewpoints and appearance changes. Furthermore, the re-identification method should be computationally efficient, so that it can be performed in every frame to bound the tracking drift. To re-identify global features with computational efficiency, we first build a spatial-temporal sensitive (STS) sub-global map, then re-identify features with pose guidance in a long time span. Finally, the re-identified features provide augmented visual measurements and are incorporated into local and global bundle adjust (BA) optimization modules for accurate pose estimation. Fig. 2 shows our proposed feature ReID process. Take the public EuRoC DB V2\_03 sequence for example, a red dot indexed by 668 in the simulated camera motion represents a map point. The point is reconstructed in the 185$^{th}$ frame and tracks to the 244$^{th}$ frame. The feature track fails in the 245$^{th}$ frame because it is approaching to the image boundary and can not be tracked correctly. With our feature ReID method, the feature is successfully re-identified in the 776$^{th}$ frame. The sight-of-view angle between the detected frame and the re-identified frame is 49.6$^{\circ}$ and the spatial distance between the two frames is about 3.3 meters. See the supplementary material for more details.
\begin{figure}[t]
\begin{center}
\includegraphics[width=1.0\linewidth]{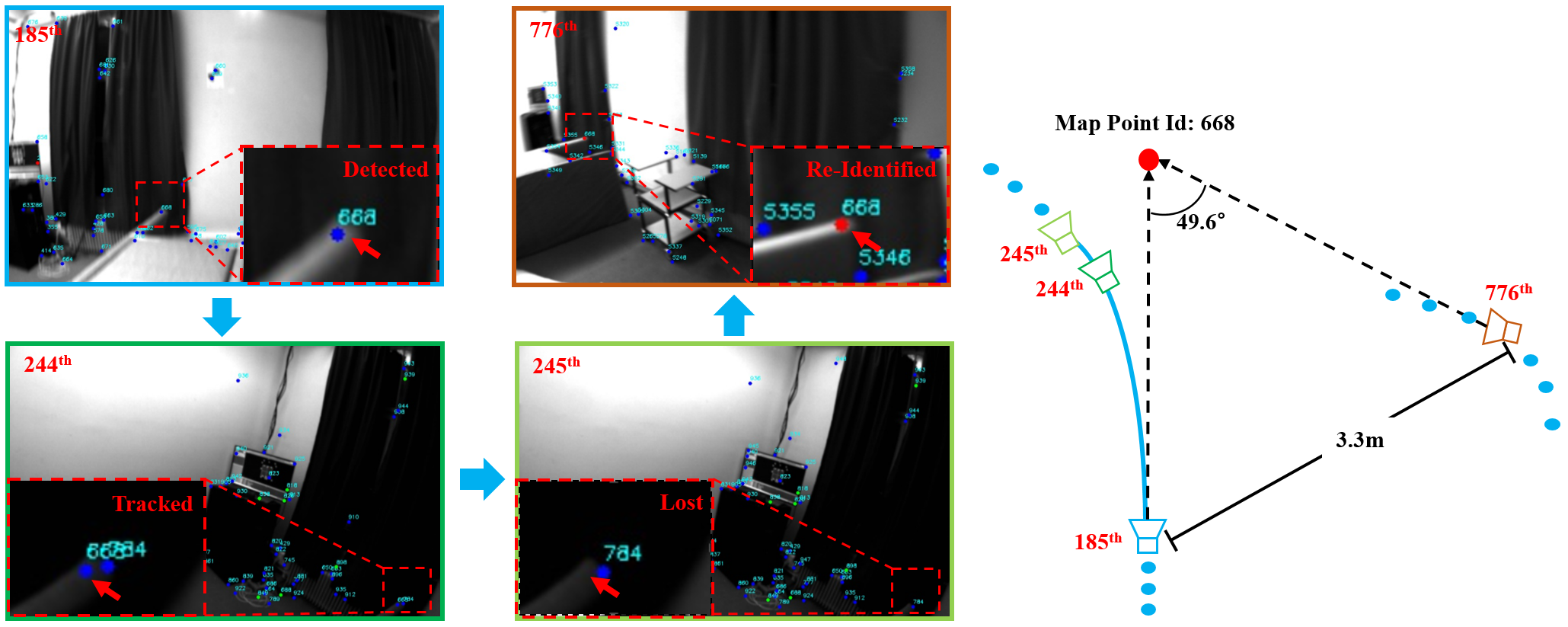}
\end{center}
\caption{Feature ReID in SLAM system. The left two column images show feature ReID process. The feature pointed by the red arrow is firstly detected in the 185$^{th}$ frame and continuously tracked to the 244$^{th}$ frame, then the feature is lost in the 245$^{th}$ frame. In the 776$^{th}$ frame, the feature is re-identified. The right graph is a part of camera tracking trajectory and positions.}
\label{fig:net}
\end{figure}

In summary, our main contributions are as follows:

\begin{itemize}

\item We propose to reconstruct a spatial-temporal sensitive (STS) sub-global map to re-identify features for every frame with high efficiency. The STS sub-global map is reconstructed from the map points in the early keyframes which satisfy multi-view geometry constraint with the current frame.
\item A pose guided feature matching method is proposed to establish feature matching. By using a fused IMU-aided, geometry and appearance consistent method, our solution enables to incorporate efficient visual measurements into energy function to bound camera drift.
\item With the combination of our proposed feature ReID method with baseline method \cite{VISLAM1}, the absolute translation error (ATE) is 3.2cm and 1.1cm on two public DBs: EuRoC \cite{c7} and TUM VI \cite{c8}, and achieves 67.3\% and 87.5\% error reduction respectively.

\end{itemize}

\section{RELATED WORK}

Tremendous research works on visual inertial SLAM have appeared in the last few years. These methods exploit monocular/stereo vision and IMU information to track camera pose and map the environment at the same time. Graph optimization based tightly-coupled visual inertial SLAM methods jointly optimize camera and IMU measurements with BA or incremental bundle adjust (IBA) \cite{c18} method from the raw measurements and achieve the state-of-the-art performance in recent years. This work focuses on graph optimization based SLAM method and closely relates to the following research topics: feature matching, pose graph, loop closure and re-localization. The following sections elaborate on the related works.

\subsection{Feature matching}

Feature matching is an important part in visual SLAM system and it directly impacts both localization accuracy and robustness. There are two types of feature matching methods in the state-of-the-art SLAM system. One is pose-free feature matching which establishes 2D-2D match by either KNN search or optical flow tracking. Liu. et al. \cite{VISLAM1} and Qin. et al. \cite{VIO2} use KLT tracker \cite{c4} and build 2D-2D correspondence. While recent optical flow methods based on deep learning are also popular \cite{DLFLOW1,DLFLOW2,DLFLOW3,DLFLOW4,DLFLOW5}. SEVIS \cite{VISLAM2} employs keyframe-aided 2D-2D feature matching to find reliable correspondences between current 2D visual measurements and 3D map features. Recent deep learning based methods \cite{c10}, \cite{c27}, \cite{c28} focus on learning better sparse detectors and local descriptors from data using convolutional neural networks (CNNs). With the deep features, Sarlin. et al. \cite{c26} learn feature matching and outlier filtering by solving a partial assignment problem. The other type feature matching is pose-assisted 3D-2D matching. The initial pose estimation of query image is from motion model or other sensors measurements, i.e. IMU integration \cite{c23}. The extracted 2D features in query image establish 2D-3D matches with 3D points in the local map attempt to estimate a 6-DoF pose with a PnP \cite{c29} geometric consistency check within a RANSAC scheme \cite{c30}.

\subsection{Pose graph}

Pose graph is defined by co-visibility and two poses are connected to each other if they share enough common features \cite{c25}. Co-visibility graph in \cite{VISLAM6} is represented as an undirected weighted graph. Each node is a keyframe and an edge between two keyframes exists if they share enough observations of the same map points. Our proposed STS sub-global map is related to the co-visibility graph which both are reconstructed/built from keyframes or its map points. However, the co-visibility graph is reconstructed by the local neighboring keyframes who have strong co-visibility relationship with the current frame, which restricts its capability to identify frames in a long time span.

\subsection{Loop closure and Re-localization}

Loop closure detection in SLAM is a key technique and is used to reduce camera drift when the camera goes back to the previous explored circumstance. Re-localization has the ability to recover from camera tracking failure in a SLAM system with a previously built map. The two tasks are relevant and many techniques are proposed in the literature working on the tasks. The approach based on bag of words (BoW), such as DBoW2 \cite{c12}, is the most popular for real-time visual SLAM systems \cite{VIO2, VISLAM2, VISLAM6}. Lynen et al. \cite{c15} propose a visual localization for large scale scene and server-side deployment. Recently, ConvNet-based approaches have risen in popularity. Merril. et al. \cite{c13} learn an auto-encoder from the common HOG descriptors for the whole image. Kuse et al. \cite{c16} learn the whole-image-descriptor in a weakly supervised manner based on NetVLAD \cite{c14}.

\section{OUR METHOD}

The structure of our proposed visual-inertial SLAM system pipeline is shown in Fig. 3. The input streams are from two different kinds of sensors: IMU and stereo camera. In the frond-end, local features are detected, tracked and IMU instant measurements are pre-integrated. In the back-end, local BA optimizes temporal latest frames in a local sliding window. Global BA optimizes all the keyframes and map points to maintain a global consistent map. Our proposed feature ReID module builds the constraints between the front-end and the back-end in SLAM system to further refine the map and control camera drift.
\begin{figure}[t]
\begin{center}
\includegraphics[width=1.0\linewidth]{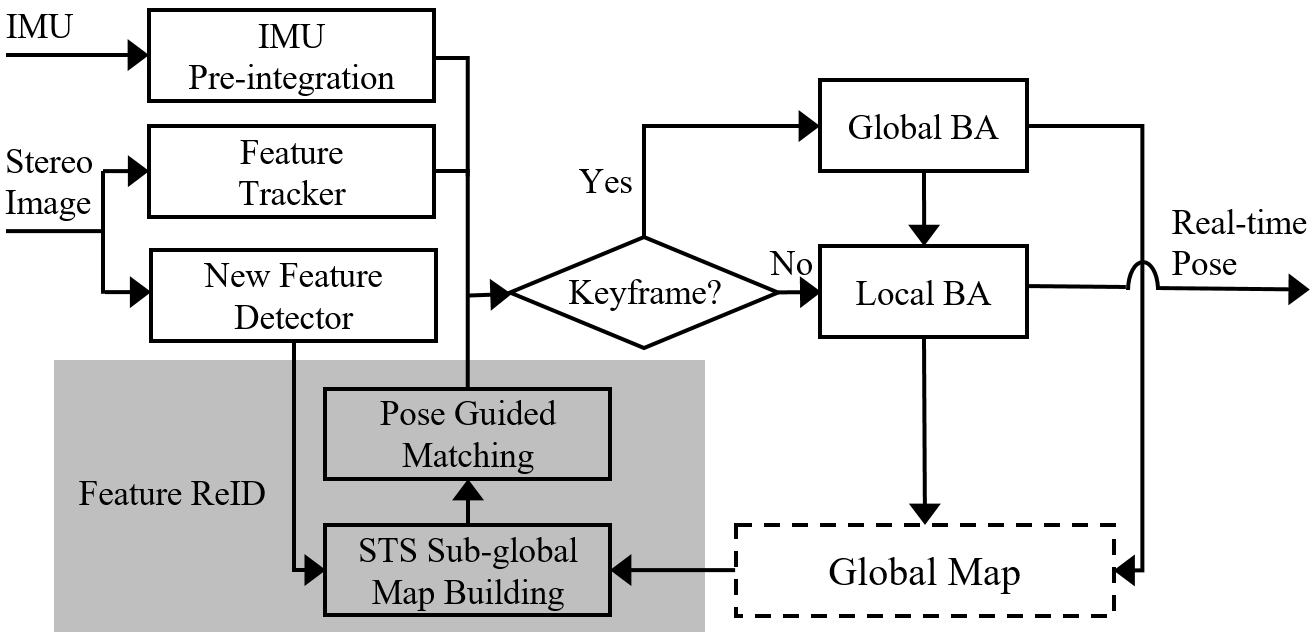}
\end{center}
\caption{Our visual-inertial SLAM method pipeline.}
\label{fig:net}
\end{figure}
\subsection{Visual Inertial SLAM}
In our method, stereo camera are equipped to guarantee the system launch under a true scale motion and map. Similar to most of visual inertial SLAM methods, our objective is to estimate the unknown camera-rate state which includes camera pose, velocity, IMU bias as well as the position of 3D map points in the environment.

Suppose camera pose is described by $\boldsymbol{T}=(\boldsymbol{R},\boldsymbol{p})$. For each 3D point $\boldsymbol{X}_j$, it is observed from multiple image frames and its corresponding 2D measurement is denoted by $\boldsymbol{z}_{ij}$ in $ith$ frame, then the visual constraint is represented by re-projection error $\boldsymbol{E}_{ij}^{vis}(\boldsymbol{T}_i,\boldsymbol{X}_j )= \pi{(\boldsymbol{T}_i}\circ{\boldsymbol{X}_j)}-\boldsymbol{z}_{ij}$, $\pi$ is a transform from the camera coordinate system to the image coordinate system. Assuming $\boldsymbol{X}_j$ is reconstructed from $s_jth$ frame and is parameterized by its inverse depth ${\rho}_j$, then $\boldsymbol{X}_j=\boldsymbol{T}_{s_j}^{-1}\circ{\frac{1}{{\rho}_j}}\boldsymbol{z}_{s_{j}j}$. IMU measurements are also important to provide relative motion constraint and are usually pre-integrated \cite{c23} to estimate IMU state $(\boldsymbol{T},\boldsymbol{M})$, where $\boldsymbol{M}=(\boldsymbol{v},\boldsymbol{b})$ means velocity and bias respectively. Following tightly coupled visual inertial localization methods \cite{VIO2}, \cite{VISLAM1}, local sliding window based nonlinear optimization framework processes visual and inertial measurements and the cost function $\boldsymbol{E}_L$ is defined as:
\begin{equation}\label{local visual}
\begin{aligned}
\boldsymbol{E}_L=\mathop{\arg\min}_{\boldsymbol{T}_{i},\boldsymbol{M}_{i},\rho_{j}}\sum_{i=t_{0}}^{t_n}\sum_{j\in{\boldsymbol{V}_{i}}}\Vert{\boldsymbol{E}_{ij}^{vis}(\boldsymbol{T}_{i}, \boldsymbol{T}_{s_{j}}, \rho_{j})}\Vert\\+\Vert{\boldsymbol{E}_{t_0}^{prior}(\boldsymbol{M}_{t_{0}},\boldsymbol{T}_{t_{0}})}\Vert\\+\sum_{i=t_{0}}^{t_n-1}\Vert{\boldsymbol{E}_{i,i+1}^{imu}(\boldsymbol{M}_{i},\boldsymbol{M}_{i+1},\boldsymbol{T}_{i},\boldsymbol{T}_{i+1})}\Vert
\end{aligned}
\end{equation}
where $t_0$ is the first frame and there are $t_n-t_0+1$ frames in the sliding window. When the oldest frame moves out of the sliding window, its corresponding visual and inertial measurements turn into a prior in local BA. If the frame is a keyframe, the prior becomes relative constraints, which is added to global BA with visual and inertial measurements of the keyframe. Global BA runs in parallel to local BA at a relatively lower frequency. The cost function $E_G$ is defined as:
\begin{equation}\label{global visual}
\begin{aligned}
\boldsymbol{E}_G=\mathop{\arg\min}_{\boldsymbol{T}_{i},\boldsymbol{M}_{i},\rho_{j}}\sum_{i=k_{1}}^{k_m}\sum_{j\in{\boldsymbol{V}_{i}}}\Vert{\boldsymbol{E}_{ij}^{vis}(\boldsymbol{T}_{i}, \boldsymbol{T}_{s_{j}}, \rho_{j})}\Vert\\+\sum_{i}\Vert{\boldsymbol{E}_{i}^{rel}(\{{\boldsymbol{T}_{k\in{\boldsymbol{\mathcal{L}}_i}}}\})}\Vert\\+\sum_{i=k_{1}}^{k_m-1}\Vert{\boldsymbol{E}_{i,i+1}^{imu}(\boldsymbol{M}_{i},\boldsymbol{M}_{i+1},\boldsymbol{T}_{i},\boldsymbol{T}_{i+1})}\Vert
\end{aligned}
\end{equation}
where $k_1,k_2,...,k_m$ are keyframe indexes. For IMU term $\boldsymbol{E}_{i,i+1}^{imu}$, prior term $\boldsymbol{E}_{t_0}^{prior}$ and relative constraint $\boldsymbol{E}_{i}^{rel}$, please refer to \cite{VISLAM1} and \cite{c23} for details. Global BA optimizes all keyframes and map points to maintain a global consistent map. For the state-of-the-art SLAM methods, the visual constraints in (1) always heavily rely on feature tracking or matching in temporal neighborhood and cannot build efficient visual measurements. It also means that the state-of-the-art SLAM methods ignore inter-frame feature matching in long-term discontinuous time which leads to larger accumulative error in the long run. To reduce the error and get drift-less camera pose, in this paper, we propose feature ReID method to build more reliable visual constraints in a long time interval.

\subsection{Feature ReID}

To re-identify features at each frame, an efficient STS sub-global map and pose guided feature matching are proposed in this part. For computational efficiency, only new detected features in the current frame are re-identified. The process terminates when all points in the STS sub-global map are examined or a certain number of top ranked features are successfully re-identified.

\textbf{STS sub-global map}
In SLAM system, the global map is a map of BA-optimized consistent states and the map size gradually increases when a new scene is explored. Feature ReID on the whole global map is infeasible due to unbounded map complexity and computational cost. On the other hand, not all keyframes associated with the global map contribute equally to BA optimization. For example, keyframes which are adjacent to the current frame in temporal domain are less important to provide novel constraint; for keyframes which are spatially overlapped with the current frame, the keyframes with old timestamps are superior than the keyframes with new timestamps since ReID from old map point provides a longer time span of feature track (not necessarily continuous) and less drift which is better. So we propose to construct a STS sub-global map in consideration of both spatial and temporal aspects.

Given a set of all keyframes in the global map, these keyframes are sorted in temporal increasing order.
For each keyframe, there are a number of associated 3D map points which are visible from its own viewpoint.
We compute these 3D map points spatial distribution which approximately represents the camera view zone.
In the same way, we compute the current frame view zone, and calculate the spatially overlapped area with these keyframe view zones. If the area is non-null, the keyframe is one of the candidates for the STS sub-global map.
Then we take temporal information into consideration. Candidate keyframes with older timestamps have top prior in the STS sub-global map. Temporal neighboring keyframes which have more spatial overlapped areas are not in our scope. To prevent redundant 3D map points in the STS sub-global map and bound the complexity, we experimentally define a maximum of the STS sub-global map size $T_{map}$. (e.g. $T_{map}$ = 1000)

Fig.4 simulates the STS sub-global map reconstruction process in a bird view.
Suppose $\psi=\{F_{k_i},F_{k_{i+1}},...,F_{k_{i+m}},...,F_{rk}\}$ is  a set of all keyframes and $F_{c}$ is the current frame.
The spatial distribution of the keyframe $F_{k_i}$ is represented by a 3D cube $[\boldsymbol{C}_{k_i}^{-},\boldsymbol{C}_{k_i}^{+}]=[\boldsymbol{U}_{k_i}-\sqrt{\boldsymbol{S}_{k_i}},\boldsymbol{U}_{k_i}+\sqrt{\boldsymbol{S}_{k_i}}]$ and shown with a dashed rectangle in a bird view, where $\boldsymbol{U}_{k_i}$ and $\boldsymbol{S}_{k_i}$ are 3D points mean and variance respectively.
For simplicity, the spatial distribution of the current frame $F_c$ is replaced by its reference keyframe $F_{rk}$.
If $[\boldsymbol{C}_{k_i}^{-},\boldsymbol{C}_{k_i}^{+}]\cap{[\boldsymbol{C}_{kf}^{-},\boldsymbol{C}_{kf}^{+}]\neq{\varnothing}}$, then the keyframe $F_{k_i}$ has common view zone with the current frame $F_c$.
As the shaded area shows in Fig.4, the keyframes $F_{k_i},F_{k_{i+1}},F_{k_{i+m+1}},F_{k_{i+m+2}},F_{k_{i+m+3}}$ have common view zone with the reference keyframe $F_{rk}$ and are considered as spatial overlapped keyframes (SOKF) of the current frame $F_c$.
The STS sub-global map points are from SOKFs with older timestamps, i.e. $F_{k_i},F_{k_{i+1}},F_{k_{i+m+1}}$, which are denoted by dashed circles without red cross.
\begin{figure}[t]
\begin{center}
\includegraphics[width=1.0\linewidth]{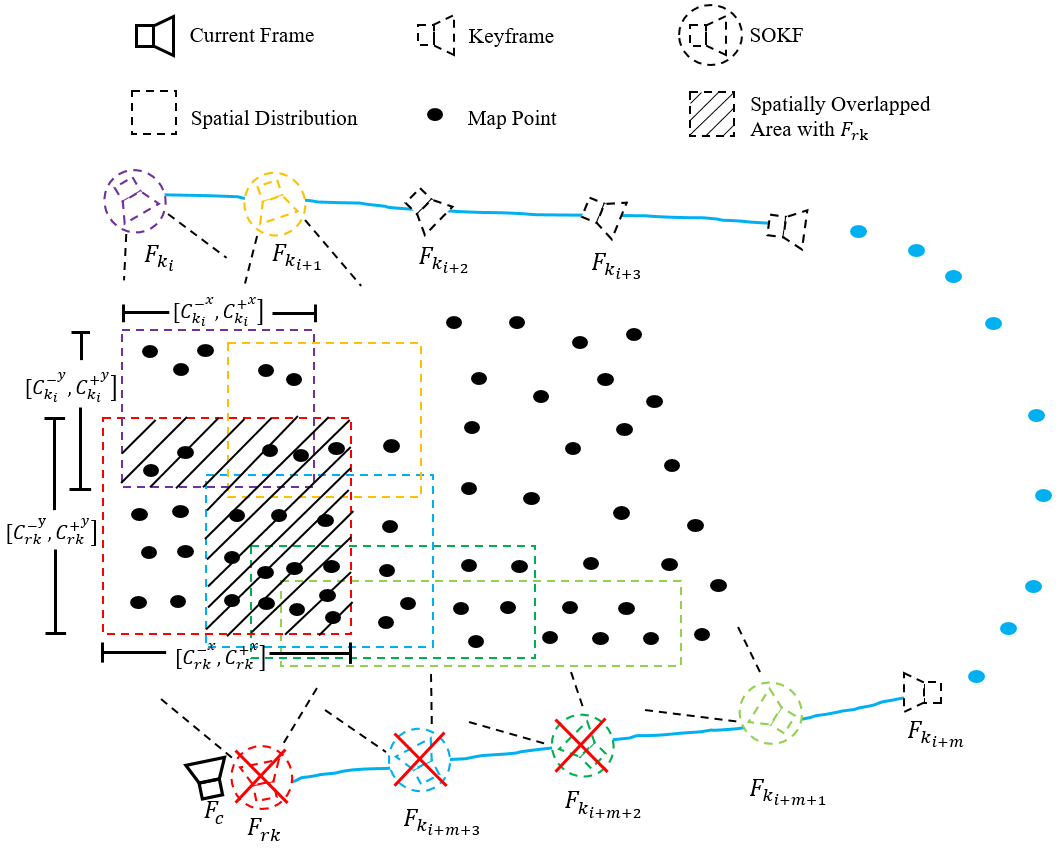}
\end{center}
\caption{The STS sub-global map reconstruction process in a bird view.
The camera tracks in a clockwise direction. The spatial distribution of each keyframe is visualized by a dotted rectangle.
Keyframes $F_{k_i},F_{k_{i+1}},F_{k_{i+m+1}},F_{k_{i+m+2}},F_{k_{i+m+3}}$ have spatially overlap (shaded area) with the reference keyframe $F_{rk}$ of the current frame $F_{c}$ and they are considered as SOKFs of the current frame.
The STS sub-global map points are from early SOKFs, i.e. $F_{k_i},F_{k_{i+1}},F_{k_{i+m+1}}$.}
\label{fig:net}
\end{figure}

\textbf{Pose guided matching}
Based on the reconstructed STS sub-global map, the new detected features in the current frame are re-identified with a pose guided feature matching method. In comparison with tracking the local map module in \cite{VISLAM6}, our method additionally utilizes IMU pre-integration, geometry filtering and warping modules to guarantee accurate and robust matching.

We exploit IMU aided camera pose prediction \cite{c23} to get a relative accurate pose prediction for finding the correspondence between 3D map points and 2D features. IMU measurements $\{(\tilde{\boldsymbol{\omega}}_k,\tilde{\boldsymbol{\alpha}}_k)\mid{k=k_1,...,k_n}\}$ in $[i,i+1]$ time intervals combining with optimal latest camera pose $(\boldsymbol{R}_i,\boldsymbol{p}_i)$, IMU bias $(\boldsymbol{b}_k^g,\boldsymbol{b}_k^a)$, velocity $\boldsymbol{v}_i$ and noise $(\boldsymbol{\eta}{_k^g},\boldsymbol{\eta}{_k^a})$ together are used to predict the current camera pose $(\boldsymbol{R}_{i+1},\boldsymbol{p}_{i+1})$ and velocity $\boldsymbol{v}_{i+1}$ via
\begin{equation}
\begin{aligned}
&\boldsymbol{R}_{i+1}=\boldsymbol{R}_i\prod_{k=k_1}^{k_n}\exp((\tilde{\boldsymbol{\omega}}_k-\boldsymbol{b}^g_k-\boldsymbol{\eta}^{gd}_k)\Delta{t}) \\
&\boldsymbol{v}_{i+1}=\boldsymbol{v}_i+\boldsymbol{g}\Delta{t_{i,i+1}}+\sum_{k=k_1}^{k_n}\boldsymbol{R}_k(\tilde{\boldsymbol{\alpha}}_k-\boldsymbol{b}^a_k-\boldsymbol{\eta}^{ad}_k)\Delta{t} \\
&\boldsymbol{p}_{i+1}=\boldsymbol{p}_i\!+\!\sum_{k=k_1}^{k_n}[\boldsymbol{v}_k\Delta{t}\!+\!\frac{1}{2}\boldsymbol{g}\Delta{t}^2\!+\!\frac{1}{2}\boldsymbol{R}_k(\tilde{\boldsymbol{\alpha}}_k\!-\!\boldsymbol{b}^a_k\!-\!\boldsymbol{\eta}^{ad}_k)\Delta{t}^2]. \\
\end{aligned}
\end{equation}
With the reliable camera pose prediction in $(i+1)th$ moment, the STS sub-global map points are projected onto the current frame to re-identify features.

Then, each keyframe in the STS sub-global map is checked with geometry consistent to avoid mismatch. The relative pose of each keyframe and the current frame is computed and then the map points of each keyframe are re-projected onto the current frame. The keyframe is considered as drift-less if most of the projected features are spatially closed to the detected feature points of the current frame, otherwise the map points are outliers.

Finally, to deal with larger viewpoints and scale changes of feature matching, we first warp the image patches with assist of predicted pose, then compute ORB \cite{c24} descriptor and appearance distance. The feature is successfully re-identified if the minimum distance is smaller than a given threshold $T_{dist}$. (e.g. $T_{dist}$ = 50)


\subsection{Visual Constraint Augment}

Once the features are correctly re-identified, the measurements are firstly constrained in a temporally latest sliding window to be optimized in local BA. Suppose $\boldsymbol{AugV}_i$ is the correctly re-identified map point index set for $ith$ frame, $\forall{j}\in{\boldsymbol{AugV}_i}$, map point $\rho_j$ and its measurements $\boldsymbol{z}_{ij}$ are constrained by the re-projection error
\begin{equation}\label{New local visual}
\begin{aligned}
\boldsymbol{E}_L^{'}=\mathop{\arg\min}_{\{\boldsymbol{T}_{i},\rho_{j}\mid{s_j{\notin}}{[t_0,t_n]}\}}\sum_{i=t_0}^{t_n}\sum_{j\in{\boldsymbol{AugV}_i}}\Vert{\boldsymbol{E}_{ij}^{vis}(\boldsymbol{T}_{i}, \boldsymbol{T}_{s_{j}}, \rho_{j})}\Vert.
\end{aligned}
\end{equation}
The new visual constraints $\boldsymbol{E}_L^{'}$, together with $\boldsymbol{E}_L$ in (1) is optimized in the sliding window. In (4), the camera states $\boldsymbol{T}_i$ and the reconstructed points $\rho_j$ during $[t_0,t_n]$ are optimized. For map points $\rho_j$, if their reconstructed frame $s_j$ is outside of the sliding window, i.e. $s_j{\notin}[t_0,t_n]$, the points $\rho_j$ are not optimized. These map points anchor the sliding window and enable that the camera poses and local points in the window are consistent with the global map.

When the oldest frame in the sliding window is a keyframe and it consists of newly re-identified features, the visual measurements in global BA cost function are also augmented and denoted by $\boldsymbol{E}_G^{'}$ in (5). Together with $\boldsymbol{E}_G$ in (2), the cost is optimized
\begin{equation}\label{New gobal visual}
\begin{aligned}
\boldsymbol{E}_G^{'}=\mathop{\arg\min}_{\{\boldsymbol{T}_{i},\rho_{j}\}}\sum_{i=k_1}^{k_m}\sum_{j\in{\boldsymbol{AugV}_i}}\Vert{\boldsymbol{E}_{ij}^{vis}(\boldsymbol{T}_{i}, \boldsymbol{T}_{s_{j}}, \rho_{j})}\Vert
\end{aligned}
\end{equation}
where $k_1,k_2,...,k_m$ are all keyframes indexes. Different from local BA, all points in $\boldsymbol{AugV}_i$ are updated in global BA which help maintain a global consistent map.

\subsection{Feature ReID Verification}

To verify the geometry consistency of the new re-identified features, the average re-projection error $\boldsymbol{E}^{vis}(\rho_j)$ of the map point $\rho_j$ is calculated by
\begin{equation}
\begin{aligned}
\boldsymbol{E}^{vis}(\rho_j)= \frac{1}{\mid{\boldsymbol{Z}_j}\mid}\sum_{i=t_0}^{t_n}\Vert{\boldsymbol{E}_{ij}^{vis}(\boldsymbol{T}_{i}, \boldsymbol{T}_{s_{j}}, \rho_{j})}\Vert
\end{aligned}
\end{equation}
where $\boldsymbol{Z}_j$ includes all 2D measurements of map point $\rho_j$, i.e. $\boldsymbol{z}_{s_{j}j}\in{\boldsymbol{Z}_j}$ and $\boldsymbol{z}_{ij}\in{\boldsymbol{Z}_j}$. The operator $\mid{\cdot}\mid$ calculates the set cardinality of $\boldsymbol{Z}_j$. If $\boldsymbol{E}^{vis}(\rho_j)$ is larger than a given threshold $T_{rep}$ (e.g. $T_{rep}$ = 10), then the map point $\rho_j$ is discarded, together with all its related measurements.

\section{EXPERIMENTS}
\begin{table}
\begin{center}
\caption{Quantitative Comparison of Average Time Span (ATS) and Average Tracking Length (ATL) on EuRoC and TUM VI}
\begin{tabular}{p{3.0cm}p{0.8cm}p{0.8cm}p{0.8cm}p{0.8cm}}
\hline\noalign{\smallskip}
\multirow{2}*{Methods}          &\multicolumn{2}{c}{EuRoC}                                       &\multicolumn{2}{c}{TUM VI}  \\  \cline{2-5}
~                               &\makecell[c]{ATS}               &\makecell[c]{ATL}              &\makecell[c]{ATS}             &\makecell[c]{ATL}  \\ \hline
\makecell[l]{Baseline(frames)}          &\makecell[c]{20.96}             &\makecell[c]{21.96}            &\makecell[c]{15.78}           &\makecell[c]{16.78} \\
\makecell[l]{Baseline + ReID(frames)}    &\makecell[c]{\textbf{43.28}}    &\makecell[c]{\textbf{22.73}}   &\makecell[c]{\textbf{79.11}}  &\makecell[c]{\textbf{19.56}} \\
Extension Rate                  &\makecell[c]{106\%}           &\makecell[c]{3.5\%}            &\makecell[c]{401\%}         &\makecell[c]{16.6\%} \\ \hline
\end{tabular}
\end{center}
\end{table}
\begin{table}
\begin{center}
\caption{Ablation study with different configurations. \ding{172} denotes STS sub-global map and \ding{173} denotes pose guided matching}
\begin{tabular}{p{2.0cm}p{3cm}p{2.0cm}}
\hline
DBs                      & Configuration                                                   & \makecell[c]{ATE(cm)} \\ \hline
\multirow{4}*{EuRoC}     & Baseline                                                        & \makecell[c]{9.8}             \\
                         & Baseline + \ding{172}                                           & \makecell[c]{3.7}             \\
                         & Baseline + \ding{173}                                           & \makecell[c]{9.1}             \\
                         & Baseline + \ding{172} + \ding{173}                              & \makecell[c]{\textbf{3.2}}  \\ \hline
\multirow{4}*{TUM VI}    & Baseline                                                        & \makecell[c]{8.8}             \\
                         & Baseline + \ding{172}                                           & \makecell[c]{1.3}             \\
                         & Baseline + \ding{173}                                           & \makecell[c]{6.9}             \\
                         & Baseline + \ding{172} + \ding{173}                              & \makecell[c]{\textbf{1.1}}  \\ \hline
\end{tabular}
\end{center}
\end{table}
\begin{table*}
\begin{center}
\caption{Average Time Cost Comparison of Each SLAM Module from ORB-SLAM3, Baseline and Ours on EuRoC and TUM VI DB}
\begin{tabular}{p{1.0cm}|p{2.5cm}|p{3.6cm}p{2.0cm}|p{3.6cm}p{2.0cm}}
\hline
        & \makecell[c]{ORB-SLAM3}        & \multicolumn{2}{c}{Baseline}                                & \multicolumn{2}{|c}{Our Method} \\ \cline{2-6}
DBs     & \multirowcell{2}{Tracking Thread(ms)}  & \makecell[c]{Tracking Thread(ms)}              & \multirowcell{2}{GBA Thread(ms)}         & \makecell[c]{Tracking Thread(ms)}             & \multirowcell{2}{GBA Thread(ms)} \\
        &                                    & \makecell[c]{(Front-end+LBA)}        &                                  & \makecell[c]{(Front-end+ReID+LBA)}  & \\ \hline
EuRoC   &  \makecell[c]{51.46}              & \makecell[c]{\textbf{11.25}}               & \makecell[c]{18.97}               & \makecell[c]{16.92}                       & \makecell[c]{23.13}      \\
TUM VI  &  \makecell[c]{48.13}              & \makecell[c]{\textbf{6.29}}                & \makecell[c]{7.86}                & \makecell[c]{11.27}                       & \makecell[c]{11.11}     \\ \hline
\end{tabular}
\end{center}
\end{table*}
To validate our proposed feature ReID method, we calculate average tracking length and time span and compare with the baseline method. Furthermore, localization accuracy is quantitatively compared with the state-of-the-art SLAM methods on two public DBs: EuRoC dataset \cite{c7} and TUM VI benchmark \cite{c8}.

\subsection{Public Datasets and Measurements}

\textbf{EuRoC Dataset}
The EuRoC dataset contains 11 sequences recorded by a small-scale hexacopter UAV, which is equipped with a visual-inertial (VI) sensor unit. The VI sensor unit provides WVGA stereo grayscale images with a baseline of 11 cm at a rate of 20 Hz and 200 Hz inertial data. The dataset is captured in two different scenes, one is a large machine hall which includes 5 sequences. The other one is a small-scale room which has 6 sequences. The ground truth pose is got by a motion capture system. Depending on the texture, brightness, and UAV dynamics the sequences are classified as easy, medium, and difficult.

\textbf{TUM VI Benchmark}
The TUM VI benchmark provides several types of sequences, such as rooms, corridors, magistrales, outdoors and slides. The room sequences can be representative of typical AR/VR applications, where the user moves wearing a head-mounted device in a small environment. We evaluate the state-of-the-art methods on this scene because it provides ground truth data in whole trajectories. The image resolution is 512x512 at 20Hz and IMU measures at a rate of 200Hz. OptiTrack motion capture system records accurate ground-truth poses at a high frame rate of 120 Hz. In these sequences, the cameras work with a circle motion in a small scale room.

\textbf{Measurements}
To validate feature re-identification effectiveness, we calculate time span (TS) and tracking length (TL) of each map point. $TS(\rho_j)=i-s_j$ defines the difference between the newest measurement frame index $i$ and the reconstructed frame index $s_j$. $TL(\rho_j)=\mid{\boldsymbol{Z}_j}\mid$ counts all 2D measurements $\boldsymbol{z}_{ij}\in{\boldsymbol{Z}_j}$ corresponding to a 3D point $\rho_j$. To evaluate SLAM localization accuracy, absolute translation error (ATE) in \cite{c22} is used to compare our method with the state-of-the-art methods. When testing all DBs, we use the same parameter settings.

\subsection{Ablation Study}

\textbf{TS and TL}
We compute average time span (ATS) and average tracking length (ATL) on EuRoC and TUM VI respectively. As shown in Table \uppercase\expandafter{\romannumeral1}, with our proposed feature ReID method, ATS has increased about 1 time and 4 times when comparing with the baseline method \cite{VISLAM1} on two different DBs. ATL has also averagely increased about 1 frame and 3 frames for each map point. We observe that the extension rate is different for the two DBs and it closely relates to circumstances and camera movements. Comparing with EuRoC DB, TUM VI benchmark is captured by a loopy camera motion in a relatively small indoor office and it's easier to re-identify features.

\textbf{Feature ReID Validation}
In this part, we validate feature ReID method with different configurations and compare with the baseline method. Table \uppercase\expandafter{\romannumeral2} lists ATE on the two public DBs and the results show that our feature ReID method achieves 67.3\% and 87.5\% error reduction when comparing with the baseline method \cite{VISLAM1}. Table \uppercase\expandafter{\romannumeral2} also validates that both STS sub-global map and pose guided feature matching are important components to improve localization accuracy in feature ReID. The STS sub-global map module preserves the most relevant map points with the features in the current frame and guarantees fast speed and makes correct feature matching possible. Pose guided feature matching module re-identifies more features, which can combine into BA optimization to further decrease accumulative error and improve localization accuracy.

\textbf{Time Analysis}
The experiments are carried out on a desktop PC with i7 3.4GHz CPU and 8G memory. The computational time comparison shows in Table \uppercase\expandafter{\romannumeral3}.
We notice that our feature ReID module costs additional 4ms$\sim$6ms on the two public DBs in the tracking thread. In comparison with ORB-SLAM3 \cite{VISLAM8}, both the baseline \cite{VISLAM1} and our method have a lower time cost.

\subsection{Comparison with the state-of-the-art}
\begin{table}
\begin{center}
\caption{Comparison with the State-of-the-art Methods on EuRoC}
\begin{tabular}{p{0.8cm}p{1.0cm}p{1.0cm}p{1.0cm}p{1.0cm}p{1.1cm}}
\hline\noalign{\smallskip}
EuRoC	& \makecell[c]{VINS \\w$\slash$ loop}	& \makecell[c]{SOFT-\\SLAM \\w$\slash$ loop}   & \makecell[c]{ICE-BA \\w$\slash$ loop}	   & \makecell[c]{ORB-\\SLAM3 \\w$\slash$ loop}	 &  \makecell[c]{Ours \\w$\slash$o loop} \\ \hline
MH\_01	      &  \makecell[c]{15.9}	&    \makecell[c]{\textbf{2.8}}	&   \makecell[c]{11}	& \makecell[c]{3.9}	             &   \makecell[c]{3.4} \\
MH\_02	      &  \makecell[c]{15.6}	&    \makecell[c]{4.2}	        &   \makecell[c]{8}	    & \makecell[c]{3.8}	             &   \makecell[c]{\textbf{2.6}} \\
MH\_03	      &  \makecell[c]{12.8}	&    \makecell[c]{3.8}	        &   \makecell[c]{5}	    & \makecell[c]{2.8}	             &   \makecell[c]{\textbf{2.5}} \\
MH\_04	      &  \makecell[c]{28.0}	&    \makecell[c]{9.6}	        &   \makecell[c]{13}	& \makecell[c]{\textbf{5.8}}	 &   \makecell[c]{6.2} \\
MH\_05	      &  \makecell[c]{29.1}	&    \makecell[c]{5.8}	        &   \makecell[c]{11}	& \makecell[c]{10.0}	         &   \makecell[c]{\textbf{5.3}} \\
V1\_01	      &  \makecell[c]{7.8}  &    \makecell[c]{4.2}	        &   \makecell[c]{7}	    & \makecell[c]{3.6}	             &   \makecell[c]{\textbf{3.4}} \\
V1\_02	      &  \makecell[c]{6.8}	&    \makecell[c]{3.4}	        &   \makecell[c]{8}	    & \makecell[c]{\textbf{1.1}}	 &   \makecell[c]{2.2} \\
V1\_03	      &  \makecell[c]{11.3}	&    \makecell[c]{5.7}	        &   \makecell[c]{6}	    & \makecell[c]{2.3}	             &   \makecell[c]{\textbf{2.1}} \\
V2\_01	      &  \makecell[c]{5.9}	&    \makecell[c]{7.2}	        &   \makecell[c]{6}	    & \makecell[c]{3.6}	             &   \makecell[c]{\textbf{3.0}} \\
V2\_02	      &  \makecell[c]{9.0}	&    \makecell[c]{6.9}	        &   \makecell[c]{4}	    & \makecell[c]{\textbf{1.5}}	 &   \makecell[c]{1.9} \\
V2\_03	      &  \makecell[c]{9.6}	&    \makecell[c]{17.3}         &   \makecell[c]{11}	& \makecell[c]{\textbf{2.2}}	 &   \makecell[c]{2.4} \\ \hline
Avg(cm)	      &  \makecell[c]{13.8}	&    \makecell[c]{6.4}	        &   \makecell[c]{8}	    & \makecell[c]{3.7}	             &   \makecell[c]{\textbf{3.2}} \\ \hline
\end{tabular}
\end{center}
\end{table}
\begin{table}
\begin{center}
\caption{Comparison with the State-of-the-art Methods on TUM VI}
\begin{tabular}{p{1.2cm}p{1.2cm}p{1.2cm}p{1.6cm}p{1.2cm}}
\hline\noalign{\smallskip}
TUM VI	& \makecell[c]{VINS \\w$\slash$ loop}	& \makecell[c]{ICE-BA \\w$\slash$o loop}	   & \makecell[c]{ORB-SLAM3 \\w$\slash$ loop}	 &  \makecell[c]{Ours \\w$\slash$o loop} \\ \hline
Room1	    &    \makecell[c]{7.0}	 &   \makecell[c]{13.9}	 &   \makecell[c]{\textbf{0.9}}	         &   \makecell[c]{\textbf{0.9}} \\
Room2	    &    \makecell[c]{7.0}	 &   \makecell[c]{10.2}	 &   \makecell[c]{\textbf{1.2}}	         &   \makecell[c]{\textbf{1.2}} \\
Room3	    &    \makecell[c]{11.0}	 &   \makecell[c]{10.0}	 &   \makecell[c]{1.2}	                 &   \makecell[c]{\textbf{0.9}} \\
Room4	    &    \makecell[c]{4.0}	 &   \makecell[c]{6.8}	 &   \makecell[c]{\textbf{0.8}}	         &   \makecell[c]{1.3} \\
Room5	    &    \makecell[c]{20.0}	 &   \makecell[c]{8.6}	 &   \makecell[c]{1.3}	                 &   \makecell[c]{\textbf{1.1}} \\
Room6	    &    \makecell[c]{8.0}	 &   \makecell[c]{3.3}	 &   \makecell[c]{\textbf{0.6}}	         &   \makecell[c]{1.0} \\ \hline
Avg(cm)	    &    \makecell[c]{9.5}	 &   \makecell[c]{8.8}	 &   \makecell[c]{\textbf{1.0}}	         &   \makecell[c]{1.1} \\ \hline
\end{tabular}
\end{center}
\end{table}
In this section, we compare our proposed feature ReID method with the state-of-the-art methods on EuRoC and TUM VI DBs.

In Table \uppercase\expandafter{\romannumeral4}, we show quantitatively comparison of our feature ReID method with the state-of-the-art methods on public EuRoC dataset. For SOFT-SLAM \cite{c17} and ICE-BA with loop \cite{VISLAM1}, we copy results from their published papers and run the released source code for VINS \cite{VIO2} and ORB-SLAM3 \cite{VISLAM8} three times and average them. Among the 11 sequences, our method dominates 6 sequences and is comparable with the the state-of-the-art methods for the other sequences. Notice that all methods equip with loop closure module except us.

Table \uppercase\expandafter{\romannumeral5} shows the comparison of our feature ReID method with the state-of-the-art methods on TUM VI Room benchmark. We run ICE-BA without loop \cite{VISLAM1} and ORB-SLAM3 \cite{VISLAM8} source codes three times and average them. The results of VINS \cite{VIO2} method are copied from \cite{c8}. It can be seen that the performance of our method exceeds VINS, ICE-BA and is comparable with ORB-SLAM3 with loop method.

\section{CONCLUSIONS}

In this paper, we propose a feature re-identification method to recognize new features from existing map and the experimental results have validated our method. More importantly, our feature ReID method can be effortlessly applied in all existing visual SLAM methods to improve ego-camera pose accuracy. This aspect is of great value for practical applications.




%
%
%
%
%

\end{document}